# Explaining Deep Learning Models for Age-related Gait Classification Based on Time Series Acceleration

Xiaoping Zheng, Michiel F. Reneman, Egbert Otten*, and Claudine JC. Lamoth

*Abstract*—*Objective:* **Gait analysis holds significant importance in monitoring daily health, particularly among older adults. Machine learning, notably deep learning (DL), shows promise in gait analysis. However, the inherent black-box nature of these models poses challenges for their clinical application. This study aims to enhance transparency in DL-based gait classification for aged-related gait patterns using Explainable Artificial Intelligence (XAI), such as SHapley Additive exPlanations (SHAP).** *Methods:* **A total of 244 subjects, comprising 129 adults and 115 older adults (age>65), were included. They performed a 3-minute walking task while accelerometers were affixed to the lumbar segment L3. DL models, convolutional neural network (CNN) and gated recurrent unit (GRU), were trained using 1-stride and 8-stride accelerations, respectively, to classify adult and older adult groups. SHAP was employed to explain the models' predictions.** *Results:* **CNN and GRU both achieved a satisfactory performance with an accuracy of 81.4% and 84.5%, respectively. SHAP analysis revealed that both CNN and GRU assigned higher SHAP values to the data from vertical and walking directions, particularly emphasizing data around heel contact, spanning from the terminal swing to loading response phases. Furthermore, SHAP values indicated that GRU did not treat every stride equally.** *Conclusion:* **CNN effectively used single stride data, whereas GRU focused on stride relationships for classification. Data at heel contact was crucial, indicating varying acceleration and deceleration patterns in different age groups.** *Significance:* **XAI enhances the transparency of machine learning-based decision-support systems, thereby facilitating their incorporation into clinical practice.**

*Index Terms*— **Gait analysis, Walking, Ageing, Deep learning, Machine learning, Explainable Artificially intelligence, Accelerometers.**

## I. INTRODUCTION

GAIT Gait analysis plays a significant role in monitoring the quality of life, particularly among older individuals, since maintaining mobility and independence in later years is essential [1]. Gait performance can provide insights into the control and coordination of various systems, such as neuromusculoskeletal system and nervous system. Aging, as a continuous process, is often associated with the loss of muscle mass, decreased bone density, and declining nerve function, which can result in an altered gait pattern [2].

With the development of miniaturization of sensors (e.g., accelerometers), modern movement tracking systems can provide vast amounts of reliable data about human movements [3], allowing for the diagnosis, monitoring, and rehabilitation of gait patterns in daily living environments. Given the high variability, dimensionality, non-linear interactions, and temporal dependencies of the data collected during walking, traditional statistical approaches have limited capabilities [4]. Therefore, machine learning (ML) approaches have gained importance in clinical gait analysis due to their ability to handle complex data.

Machine learning has demonstrated promising results in clinical gait classification tasks. For example, Artificial Neural Network (ANN) achieved high accuracy (90%) in classifying different age groups gait based on handcrafted gait outcomes [5]. The design and selection of handcrafted gait outcomes require expert knowledge and are laborious. Deep learning (DL) can perform gait classification based on raw sensor signals and has demonstrated superior performance [6]. A recent study compared the classification performance of recurrent neural networks (bidirectional long short-term memory) and conventional machine learning (support vector machine (SVM) and linear regression) in classifying fallers and non-fallers in patients with multiple sclerosis [7]. The results of this study indicated that the deep learning approach outperformed conventional machine learning (area under the curve: 0.88 vs. 0.79 for SVM).

However, many machine learning models suffer from a lack of transparency and interpretability due to their black-box nature [8]. It is often unclear why a specific prediction has been made, even though the mathematical principles underlying these methods are well-established and well-understood. This opacity makes it challenging for patients and clinicians to trust the models, and strongly limits their practical applications in clinical contexts. Furthermore, this lack of transparency does not comply with the requirements of the European General Data Protection Regulation (GDPR, EU 2016/679) [9], which mandates the explanation of the logic

X. Z. was supported by the China Scholarship Council-University of Groningen Scholarship under grant No.201906410084.

M. R. is with the Department of Rehabilitation Medicine, University of Groningen, University Medical Center Groningen, 9713 AV, Groningen, the Netherlands.

X. Z., E. O., and *C. L. are with the Department of Human Movement Sciences, University of Groningen, University Medical Center Groningen, 9713 AV, Groningen, the Netherlands (correspondence e-mail: c.j.c.lamoth@umcg.nl).

behind any automated decision-making process that significantly affects individuals. Apart from this, the black-box nature makes it impossible to know what the model has truly learned, consequently obstructing the potential for generating new knowledge and a better understanding of human gait movement.

To overcome these limitations, Explainable Artificial Intelligence (XAI) has gained attention in the field of medicine. XAI is an approach aimed at revealing the reasoning behind a system's predictions and decisions, which becomes even more critical when handling sensitive and personal health data. XAI can be broadly categorized into two main categories based on the stage of use: 1) ante-hoc explainability; and 2) post-hoc explainability [10]. Ante-hoc explainability refers to simple models that are interpretable by design, such as linear regression models, decision trees, k-nearest neighbour models, and Bayesian models [11]. However, it is often assumed that ante-hoc explainable models do not achieve satisfactory performance; therefore, opaque models (such as deep learning) are frequently employed [11]. This leads to post-hoc explainability approaches, which can be used to explain a previously trained model or its prediction.

Layer-wise Relevance Propagation (LRP) [12, 13], Local Interpretable Model-Agnostic Explanations (LIME) [14], and SHapley Additive exPlanations (SHAP) [15] are the popular post-hoc explainability approaches. LRP propagates relevance scores from the output layer back to the input layer to determine the relevance of each input variable to the output decision. LIME perturbs the original data to observe how it affects predictions and aims to provide interpretable and faithful explanations, but it suffers from instability. It has been reported that two very close input samples may get greatly varied explanations in a simulated setting [16]. SHAP utilizes SHapley values to represent the contribution of each input variable to a certain prediction and this approach has strong theoretical backing and ensures that the contribution of each input variable is fairly and efficiently distributed among all the input variables of the instance [17]. This efficiency property distinguishes SHAP from other approaches and suggests that it might be the only approach that provides a full and fair explanation for the prediction of a machine learning model. This property may highlight the potential advantages of SHAP values in terms of fairness and legal compliance in certain situations.

The application of XAI approaches in deep learning-based clinical gait analysis is still in early stages. One study used LRP to explain convolutional neural network (CNN) in classifying individual gait patterns based on one stride data collected by ground reaction forces and full-body joint angles [12]. In another study, CNN was used to classify the walking of healthy subjects while performing four different dual tasks. The classification was based on data from no more than two strides, derived from ground reaction force and plastic optical fiber distributed sensors. [13]. In this study, LRP was used to indicate which parts of the signal have the heaviest influence on the gait classification. Both studies applied a CNN model-based classification, which is famous for extraction of local temporal and spatial features. However, since only 2 strides of data were used as input, the long-term dependent changes in gait, which provide information about postural control ability [18], were not considered.

The present study aims to improve the transparency of DL-based gait classification, with time-series acceleration obtained during a 3-minute walking task as input. The goal is to differentiate between the gait patterns of adults and older adults. More specifically, we will: 1) employ a CNN and a gated recurrent unit (GRU) designed to learn long-term dependent features; 2) utilize the SHAP approach to indicate the importance of the input signal in classification for both models. This study can contribute to improving the transparency and interpretability of deep learning-based gait analysis and potentially lead to better clinical decision-making.

## II. Methods

The overview of data acquisition and analysis in this study is presented in Fig. 1, with CNN being used as the example. The data were collected during a 3-minute walking task (Fig. 1(a)). Strides data from all the subjects underwent preprocessing before being employed to train the CNN model which aimed to classify subjects into adult and older adult groups (Fig. 1(b)). To interpret the CNN model, the SHAP approach was applied. The SHAP values were visualized using a colour spectrum to illustrate the contributions of input data to the classification process (Fig. 1(c)). In this representation, deeper red indicates a higher contribution.

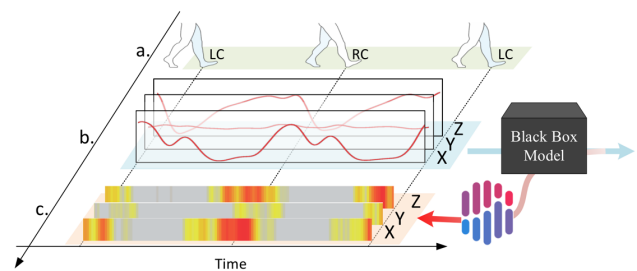

Fig. 1. Overview of data acquisition and analysis of CNN. (a): walking data collection; (b): preprocessing stride data and training CNN based on one stride data; (c) interpreting the CNN model by SHAP, deeper red colour represents a higher contribution to the classification process. CNN = convolutional neural network; LC = left contact; RC = right contact.

### A. Subjects, Equipment, and Data Collection

In this study, the dataset consisted of 386 subjects, which were derived by merging data from existing datasets[19-23]. Subjects with cognitive impairment (n=104) and those with insufficient walking data (n=38, having fewer than 10 segments of 8 consecutive strides) were excluded. The remaining subjects (n = 244) were divided into two groups: adults (ages 18-65, n=129) and older adults (ages >65, n=115). The mean ages were 38.3 (SD: 15.4) and 76.7 (SD: 5.9), respectively. During the study, subjects were instructed to walk at a comfortable speed for 3 minutes while wearing an accelerometer (iPod, Dynaport, or ActiGraph) fixed to a belt near the lumbar segment L3. When assuming a standing and upright position, the orientation of the axes was as follows: the X-axis pointed toward the ground (representing the vertical direction, V), the Y-axis faced the walking direction (indicating the anteroposterior direction, AP), and the Z-axis was perpendicular to the walking direction, extending from the patient's left to right (representing the mediolateral direction, ML). The sampling frequency was 100 Hz.

The studies were conducted between 2008 and 2022 and were approved by the Medical Ethical Committee of the University Medical Centre Groningen and the Medical Ethical Committee of the Slotervaart Hospital. All subjects provided written informed consent in accordance with the Declaration of Helsinki.

### B. Preprocessing and Data Splitting

To remove high-frequency noise, a second-order Butterworth low-pass filter with a cut-off frequency of 10 Hz was employed. The resulting filtered signal was normalized to a range of −1 to 1. A stride was defined as the period from the first left heel contact to the right contact and back to the second left heel contact. Heel contact was detected based on the peaks of both AP- and V-axis acceleration data, with the left or right foot determined by the values in the acceleration of ML direction. During the walking, the sensor sways with the movement of the body in the ML directions, and left heel contacts show higher readings in the ML direction than right contacts. To align the starting and ending timing of different strides, the data from each stride were interpolated to a uniform length of 128 samples. The segments, each with a length of 128, were employed in the CNN model, and for each subject, the initial 80 segments were chosen. In the case of the GRU model, 8 consecutive segments (stride) were merged into a singular segment comprising 1024 samples, yielding a total of 10 segments for each subject. The adult and older adult subjects were randomly and proportionally divided into training, testing, and validation sets at a ratio of 146:49:49. Their corresponding data were used as training, testing, and validation data set.

### C. Classifiers

CNN and GRU were utilized in this study because of excellent capacity of CNN for local special and temporal feature extraction and the outperformance of GRU in learning long-term dependent features [24].

The interpolated one-stride segments for CNN and interpolated eight-stride segments for GRU were organized in x-, y-, and z-axis order, to generate a single signal data with 3 channels (128*3 and 1024*3 respectively). The optimal hyperparameters for both the CNN and GRU models were tuned by Bayesian Optimization (BO) [25]. Unlike conventional techniques such as randomized search cross-validation, BO considers the prior performance of the hyperparameters and updates them to achieve better performance. This allows BO to find the global optimum with a minimum number of steps. For each model, 15 parameter combinations were tested. Detailed information about the hyperparameter space settings can be referenced in Table I,

#### TABLE I
HYPERPARAMETERS SPACE FOR CNN AND GRU

| | | Layer | CNN | GRU |
|---|---|---|---|---|
| Input | | | 128X3 | 1024X3 |
| Stack (Stack number: [1,3]) | Deep layer | Layer name | Conv1D | GRU |
| | | Unit/ Filter | [2, 768] | |
| | | Kernel size | [1, 15] | - |
| | | Activation | "ReLU" | "tanh" |
| | Batch Normalization | | - | |
| | Pooling | | - | |
| | Dropout | Rate | Nan or [0.1, 0.9] | |
| | Dense | Unit | [2, 768] | |
| | Flatten | | - | |
| | Dropout | Rate | Nan or [0.1, 0.9] | |
| Output | Dense | | 2 units and "softmax" activation | |

CNN = convolutional neural network; GRU = gate recurrent unit; ReLU = rectified linear unit; tanh = hyperbolic tangent function.

while the learning rate, which was also optimized using BO, was configured within the range of [1e-5, 1e-2].

### D. Evaluation

The assessment of classification performance was conducted using widely recognized evaluation measures such as accuracy, recall (sensitivity), precision, and F1 score (the harmonic mean of sensitivity and precision). Receiver operating characteristic (ROC) curves were generated and the area under the curve (AUC) was calculated as well.

To ensure transparency and reproducibility of the findings, we have made the project repository publicly accessible at https://github.com/xzheng93/Explainable_DL. The repository contains the source code, dataset, log files of experiments.

### E. SHAP

The SHAP approach [17] was used to explain the prediction of a signal segment $x$ in the given model $f$ (CNN or GRU) based on the SHapley values from coalitional game theory. The original input segment $x$ was mapped through the function $h_x(z')$ to get the input for the SHAP explanation $g(\bullet)$. $z' \in \{0,1\}^N$, where $N$ is the number of features (data points or sets of data point in the signal segment) of $x$ and, 0 and 1 mean the absent or present of features in $x$. Applying to the model $f: f(h_x(z'))$, the SHAP explanation [17] can be defined as:

$$g(z') = f(h_x(z')) = \emptyset_0 + \sum_{i=1}^{N} \emptyset_i z_i' \qquad (1)$$

where $\emptyset_i$ is the SHapley values of a feature $i$ in the segment $x$.

The definition of SHapley values $\emptyset_i$ is derived as follows:

$$\emptyset_i = \frac{1}{|N|!}$$
$$\cdot \sum_{\{i\}\in s \; and \; s\subseteq N} (|s|-1)! \, (|N|-|s|)! \, [f(s) - f(s-\{i\})] \qquad (2)$$

where $s$ is the segment which data features $i$ is present. $| \bullet |$ represents the length of a segment except absent features. The definition of SHapley value make it satisfies the efficiency, symmetry, dummy, and additivity properties.

The efficiency property can represent as:

$$\sum_{i \in N} \emptyset_i = f(x) \qquad (3)$$

The sum of the SHapley values of all separated features equals to the value of the coalition of all the features (the whole signal segment). Therefore, all the gain is distributed among the segment.

The symmetry property means that if the contributions of two features $i$ and $j$ are equal, they will contribute equally to all possible coalitions. And can be represented as, if $\emptyset_i = \emptyset_i$, then

$$f(s \cup \{i\}) = f(s \cup \{j\}) \qquad (4)$$

where $s \subseteq N$ and $\{i, j\} \notin s$.

The dummy property means if a feature $i$ does not change the predicted value:

$$f(s \cup \{i\}) = f(s) \qquad (5)$$

, its SHapley value $\emptyset_i$ equals to 0.

Regarding the additivity property, if a coalition game is combined by two gain functions $f'$ and $f''$, the SHapley values are additive:

$$\emptyset_i(f' + f'') = \emptyset_i(f') + \emptyset_i(f'') \qquad (6)$$

The SHapley value is built based on a solid theory. The properties of SHapley value give the explanation a reasonable foundation and distinguish the SHAP from other methods such as LIME [26]. The SHAP explanation might be the only legally compliant method to meet the law requirement [26].

## III. RESULTS

The optimal hyperparameters and architecture for both CNN and GRU were determined through BO, and the results are presented in Fig. 2 and Fig. 3. In the case of CNN, the training dataset comprised 11680 (146*80) one-stride data (segments), the testing dataset included 3920 (49*80) segments, and the validation dataset contained 3920 (49*80) one-stride segments. For GRU, the training dataset consisted of 1460 eight-stride segments (146*10), the testing dataset encompassed 490 eight-stride segments (49*10), and the validation dataset comprised 490 eight-stride segments (49*10).

The CNN architecture included three 1D convolutional layers followed by batch normalization, max-pooling, and dropout layers. The first convolutional layer consisted of 88 filters with a kernel size of 13, while the second convolutional layer comprised 336 filters with a kernel size of 5. The third convolutional layer contained only 2 filters with a kernel size of 1. The dropout rates were 0.3, 0.6, and 0 for the first, second, and third dropout layers, respectively. A dense layer with 74 units, a fully connected layer, and a dropout layer with a rate of 0.5 were also incorporated into the architecture. Finally, a softmax activation function was employed for classification. The Adam optimization was used, and the optimal learning rate of 0.0015 was discovered via BO. The model was trained for 150 epochs with early stopping based on the validation accuracy with a patience of 20 epochs.

Fig. 3 graphically illustrates the architecture and optimal hyperparameters of the proposed GRU model. The GRU architecture included three GRU layers, each followed by batch normalization, max-pooling, and dropout layers for regularization. The first GRU layer had 666 filters, the second had 438 filters, and the third had 2 filters. The three dropout layers had a rate of 0.5, 0.7, and 0, respectively. A dense layer with 676 units and a dropout layer with a rate of 0.1 were also incorporated into the architecture. The final layer of the GRU consisted of a fully connected layer followed by a softmax activation function for classification. A graphical representation of the GRU architecture is shown in Fig. 3. The Adam optimization was employed with a learning rate of 0.0003. The same training epoch setting and early stopping as CNN were used.

Based on the testing data, the classification performance of the CNN is summarized in Table II, achieving an accuracy of 81.4%, precision of 82.7%, recall of 76.3%, F1-score of 79.3%, and an AUC of 0.89. Detailed classification results for the CNN model, including the confusion matrix and ROC curve, are presented in Fig. 4. In the adult group, 85.9% of data samples were correctly classified, while 14.1% were incorrectly classified as older adults. In the older adult group, 76.3% of the samples were correctly classified, while 23.7% were incorrectly classified as adults.

Table III displays the classification performance of the GRU, with further details provided in Fig. 5. The GRU model achieved an accuracy of 84.5%, precision of 79.4%, recall of 90.4%, F1-score of 84.6%, and an AUC of 0.94. The confusion matrix in Fig. 5(a) illustrates that 79.2% of adults and 90.4% of older adults were correctly classified.

After the evaluation, the mean absolute SHAP values for the testing data of both CNN and GRU models were computed and visualized in Fig. 6. In this representation, a deeper red colour signifies a greater contribution to the classification process.

In Fig. 6(a), an abundance of red colour is observed in the V and AP directions, particularly around heel contact. A similar pattern is evident in Fig. 6(b), with a prevalence of red colour in the V and AP directions, centered around heel contact event. Notably, Fig. 6(b) reveals that not all gait cycles are equally significant, as some exhibit a higher degree of red colour, indicating greater importance in the classification process.

TABLE II
THE PERFORMANCE METRICS OF CNN

|  | Accuracy | Precision | Recall | F1-score | AUC |
|---|---|---|---|---|---|
| CNN | 81.4% | 82.7% | 76.3% | 79.3% | 0.89 |

CNN = convolutional neural network; ACU = area under the curve.

TABLE III
THE PERFORMANCE METRICS OF GRU

|  | Accuracy | Precision | Recall | F1-score | AUC |
|---|---|---|---|---|---|
| GRU | 84.5% | 79.4% | 90.4% | 84.6% | 0.94 |

GRU = gate recurrent unit layer; ACU = area under the curve.

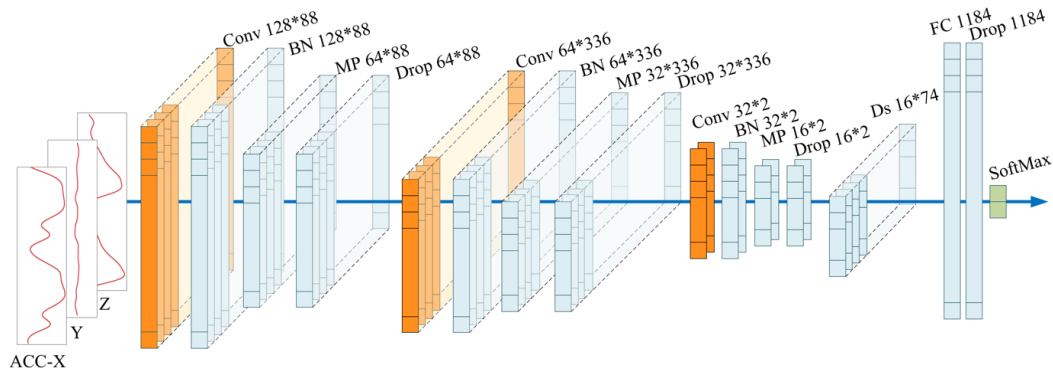

Fig. 2. The architecture and optimal hyperparameter of CNN. ACC = acceleration; CNN = convolutional neural network; Conv = 1-dimension convolutional layer (in orange); BN = batch normalization layer; MP = max-pooling layer; Drop = dropout layer; Ds = dense layer; SoftMax = softmax activation (in green).

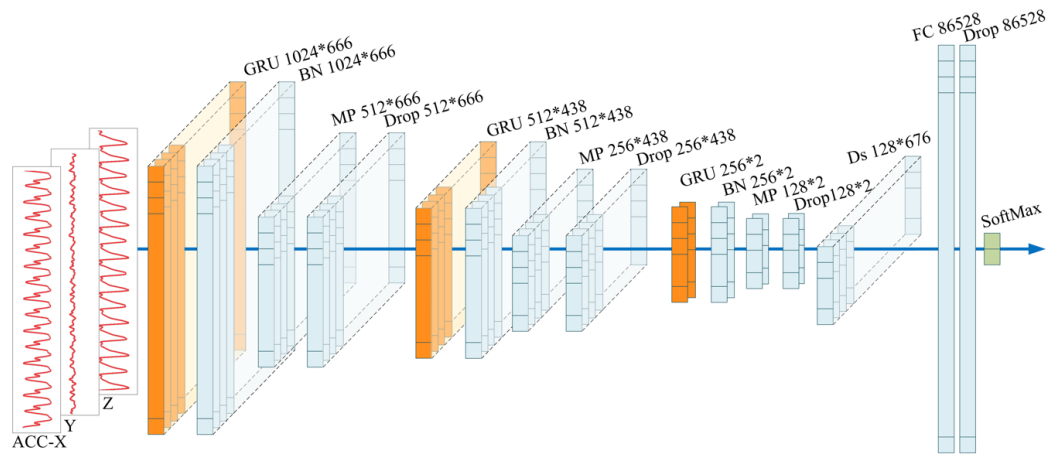

Fig. 3. The architecture and optimal hyperparameter of GRU. ACC = acceleration; GRU = gate recurrent unit layer (in orange); BN = batch normalization layer; MP = max-pooling layer; Drop = dropout layer; Ds = dense layer; SoftMax = softmax activation (in green).

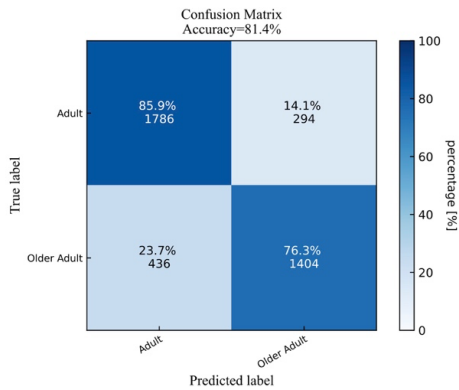

(a) Confusion matrix

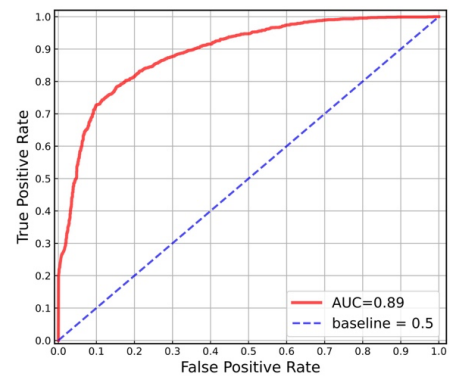

(b) Receiver operating characteristic curve

Figure 4. (a) Confusion matrix and (b) Receiver operating characteristic curve for CNN. CNN = convolutional neural network; ACU = area under the curve.

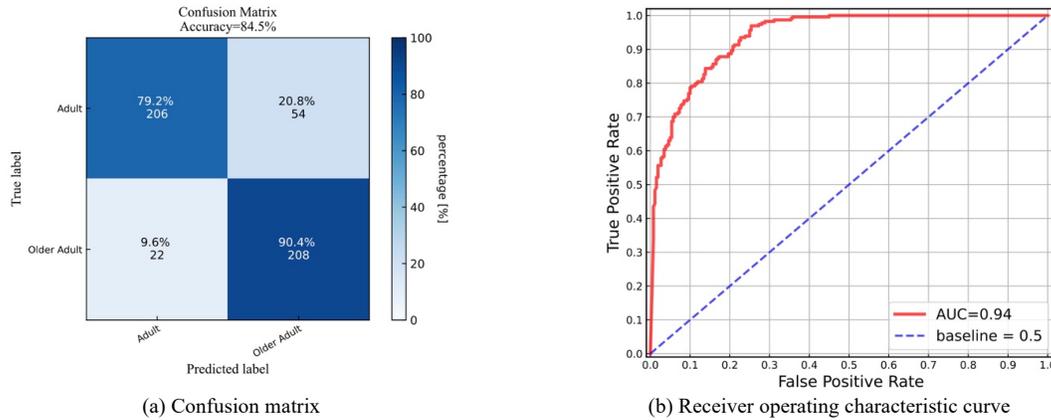

(a) Confusion matrix  (b) Receiver operating characteristic curve

Fig. 5. (a) Confusion matrix and (b) Receiver operating characteristic curve for GRU. GRU = gate recurrent unit layer; ACU = area under the curve.

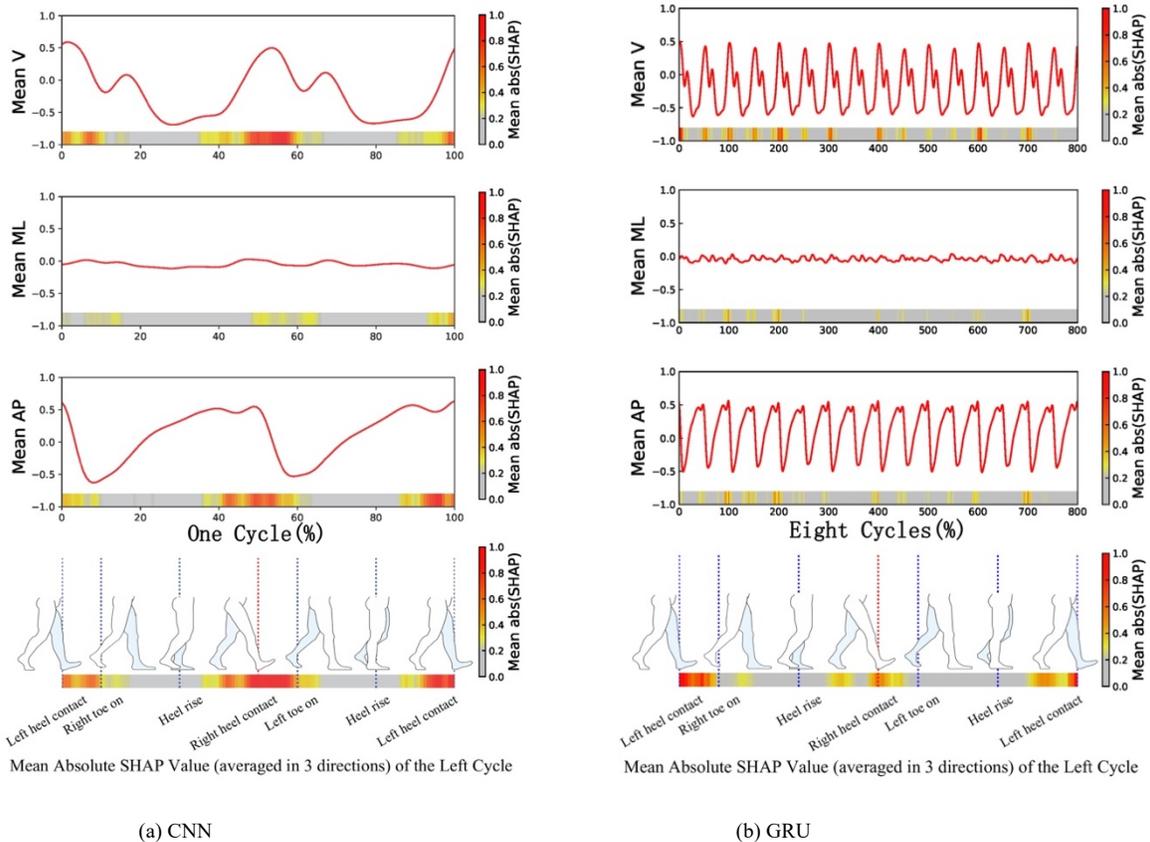

(a) CNN  (b) GRU

Fig. 6. SHAP values results of (a) CNN and (b) GRU. The acceleration data in each panel were normalized for both amplitude (ranging from -1 to 1) and time (using left heel contact as the reference point). CNN = convolutional neural network; GRU = gate recurrent unit; V = vertical direction; AP = anteroposterior direction; ML = mediolateral direction; SHAP = SHapley Additive exPlanations.

## IV. DISCUSSION

The primary aim of this study is to increase the transparency of non-linear DL models in gait analysis. For this purpose, state-of-the-art DL models, specifically CNN and GRU, were explained by using a cutting-edge XAI approach (SHAP). These models are applied to the task of classifying individuals into two distinct groups: adults and older adults, based on time series acceleration data collected during 3 minutes walking. The results indicate that CNN model achieved satisfied classification performance, with an accuracy of 81.4% and an AUC of 0.89, despite being trained on data from one stride. The GRU model exhibited promising classification capabilities, achieving an accuracy of 84.5% and an AUC of 0.94, utilizing eight-stride data. To attain an understanding and interpretation of the proposed DL models in the context of gait classification, the SHAP approach was employed. The SHAP values shed light on the models' decision-making processes, revealing a predominant reliance on acceleration data from the AP and V directions, rather than the ML direction, for the classification task. Specifically, data surrounding gait events such as heel contact in the AP and V directions emerged as the most influential inputs contributing to the differentiation between adults and older adults.

In this study, CNN and GRU were employed. CNN is renowned for its exceptional capacity to extract local spatial-temporal features. It has the potential to capture time-independent gait features, such as root mean square (indicative of gait intensity), rhythm (reflecting the proportion of stance and swing phases), and harmonic index (measuring the smoothness of the acceleration curve). These features have previously been successfully utilized in characterizing age-related gait differences in other studies [5, 27]. The promising accuracy achieved by CNN suggests that even a single stride data contains rich information that can effectively distinguish age-related gait patterns. The SHAP values underscore that data corresponding to heel contact event play an important role in this discrimination. On the other hand, GRU is designed to capture both short- and long-term dependent features. It may learn time-dependent gait features that can reflect the intricate relationships and subtle differences between gait cycles. Time-dependent gait features, including regularity, variability, local stability (as measured by the largest Lyapunov exponent), gait symmetry (using the symmetry index), and complexity (evaluated through sample entropy), are crucial in age-related gait classification. These features offer insights into changes in postural control that arise due to aging [28, 29]. The SHAP values presented in Fig. 6 (b) highlight that not all gait cycles were treated equally by GRU, as some gait cycles exhibit higher SHAP values. This observation suggests that GRU takes the relationships and slight variations between gait cycles into account when classifying individuals into the adult and older adult groups.

Aging is an ongoing process often accompanied by a gradual decline in balance [30]. Changes within the neuromusculoskeletal system, such as the loss of muscle fibers and reduced muscle force production, can result in diminished muscle strength and flexibility [31]. These alterations may impact functionality and contribute to reduced mobility. Furthermore, the sensory systems are critical for effective postural control, including the visual, vestibular, and proprioceptive systems, tend to deteriorate with age, further affecting one's balance [32]. However, the SHAP results from this study indicate that DL models predominantly rely on data from the AP and V directions, instead of the ML direction, which is more closely associated with balance capacity. This observation aligns with a recent study that supports our findings, emphasizing the significance of dynamic gait parameters in the AP and V directions for classifying age-related gait patterns [5]. These parameters include Root Mean Square in AP and V directions, Lyapunov Exponent in the V direction, step regularity in the V direction, Cross Entropy in both V and ML directions, and gait speed when utilizing artificial neural networks for classification [5]. Given that the study [5] also utilized only one accelerometer, it is possible that the similar results may be attributed to the limited sensitivity of a single accelerometer in detecting balance-related postural control information. An additional explanation for the limited contribution of ML direction data to the classification process could be attributed to the simplicity of the task undertaken in this study. Subjects engaged in a 3-minute walking task within a clean and well-lit hallway, walking at their preferred pace without any perturbations. Consequently, this task may not effectively capture the variations in balance capacity between adults and older adults in more challenging real-life environments, which may explain why ML-direction data did not play as a significant role in the classification process as AP and V direction data.

The colour spectrum of SHAP values in Fig. 6 not only reveals which axes contribute more to the classification process but also identifies specific gait events that play a crucial role. It shows that data spanning from the terminal swing to the loading response phase consistently yield higher SHAP values, particularly around the event of heel contact. The terminal swing is a phase before the heel contacts the ground. It involves the final preparations of the leg and foot for ground contact. For example, muscles around the ankle and knee are activated to ensure that the limb is prepared to provide the necessary stability during heel contact and loading response [33]. During this phase, the leg starts to slow down its forward swing to prevent excessive force upon heel strike. The acceleration, during this phase, in the walking direction (AP) has no obviously increase. The loading response phase starts with the initial contact of the heel with the ground and ends as the opposite limb toe-off the ground. After the heel contact, the body undergoes shock absorption and weight acceptance as the body's weight is transferred onto the stance limb [34]. During this phase, the acceleration data in the AP direction reach the maximum around the heel contact, and sharply decrease to reach the minimum around the toe-off event. The acceleration data around the terminal swing phase effectively represent how an individual prepares for deceleration, the moment of heel contact illustrates the process of deceleration, and the toe-off event signifies the initiation of acceleration [34].

The acceleration disparities observed by SHAP in these gait events/phases may indicate that adults and older adults exhibited different deceleration and acceleration patterns during walking. It can be attributed to the changes in kinematic and kinetic factors associated with aging which observed by previous study. Research has indicated that older adults tend to exhibit a reduced knee extension angle [35] and moment [36] at the point of heel contact, which can be closely linked to weaker muscles, such as the quadriceps [37, 38]. These changes may result in a reduced absorption force in the knee joint [39]. These alterations can lead to compensatory gait adjustments aimed at alleviating joint discomfort or stiffness, particularly in the hip and knee joints. Compared to adults, older adults exhibit limited capabilities in limb advancement during the push-off period. Research shows that, during the loading response phase, older adults often demonstrate reduced hip extension and moment [36, 40-42], particularly during the toe-off phase. This reduction may be indicative of decreased power in the hip extensors [41, 43, 44], which could imply weakness in swinging and kicking the lower limbs to generate forward propulsive force while walking. Furthermore, older adults tend to exhibit decreased independent movement of the subtendons [45] and reduced plantar flexor moments [42], contributing to lower propulsive power at the ankle [46]. The current study has insufficient data to examine whether these kinematic and kinetic factors are responsible for the distinct acceleration and deceleration patterns. Further studies are necessary in this regard.

XAI, such as the SHAP approach, provides explainability results based on input and model output data. Alterations in input signals can yield divergent outputs, and these changes may be influenced not only by aging but also by independent parameters, such as sensor brands. This study utilized three different types of accelerometers which have different dynamic ranges and accuracy. To minimize potential biases introduced by these independent parameters in prediction explanations, signal amplitude standardization and gait cycle normalization were used. It is important to note that while these techniques mitigate the bias, they may inadvertently remove valuable information, such as information related to gait intensity and walking speed. Although a prior study has demonstrated that gait cycle normalization has only a marginal impact on age-related gait patterns classification performance [6]. Notably, XAI provides explanations based on correlations and associations within the data rather than revealing causal relationships. Consequently, explanations offered by XAI may not always align with human intuition or domain expertise. These disparities can offer novel insights, or lead to misunderstandings or mistrust, particularly in critical domains like healthcare. Unfortunately, a ground truth for evaluating the quality of XAI explanations remains absent. Hence, the explainability results should be interpreted cautiously. Additionally, it is worth mentioning that the SHAP approach, while effective, can be computationally demanding and may not be optimally scalable for large datasets or real-time applications. These computational constraints limit their practical use.

## V. Conclusion

The present study enhances the transparency and interpretability of the proposed DL in gait analysis by incorporating the SHAP approach. The results demonstrate that CNN can accurately distinguish between adults and older adults based on data from a single stride. The key factors contributing to this classification were the accelerations around heel contact in the AP and V directions. GRU also exhibited promising classification performance, leveraging data from eight consecutive strides. The SHAP results from GRU suggest that it may capture the relationships and subtle variations between gait cycles, particularly the accelerations around heel contact in the AP and V directions. These findings imply that adults and older adults exhibit distinct acceleration and deceleration patterns during 3 minutes of walking.

This study underscores the potential of methods that enable understanding and interpreting machine learning predictions, such as SHAP, in advancing the application of machine learning in gait analysis. Consequently, XAI holds the promise of facilitating the implementation of machine learning-based decision-support systems in clinical practice.